\documentclass[conference]{IEEEtran}
\IEEEoverridecommandlockouts

\usepackage{cite}
\usepackage{amsmath,amssymb,amsfonts}
\usepackage{algorithmic}
\usepackage{graphicx}
\usepackage{textcomp}
\usepackage[table]{xcolor}
\usepackage{tabularx}
\usepackage{booktabs}
\usepackage{pgfplots}
\usepackage{pgfplots}
\usepgfplotslibrary{groupplots}
\pgfplotsset{compat=1.18}
\definecolor{PlotVit}{RGB}{91,169,219}
\definecolor{PlotDeit}{RGB}{96,174,92}
\definecolor{PlotSwin}{RGB}{214,128,153}
\newcolumntype{Y}{>{\centering\arraybackslash}X} 

\newcommand{\vitone}[1]{\cellcolor[RGB]{226,242,252}#1}
\newcommand{\vittwo}[1]{\cellcolor[RGB]{198,228,246}#1}
\newcommand{\vitthree}[1]{\cellcolor[RGB]{164,211,239}#1}
\newcommand{\vitfour}[1]{\cellcolor[RGB]{126,191,230}#1}
\newcommand{\vitbest}[1]{\cellcolor[RGB]{91,169,219}#1}

\newcommand{\deitone}[1]{\cellcolor[RGB]{230,246,229}#1}
\newcommand{\deittwo}[1]{\cellcolor[RGB]{203,234,199}#1}
\newcommand{\deitthree}[1]{\cellcolor[RGB]{171,217,164}#1}
\newcommand{\deitfour}[1]{\cellcolor[RGB]{135,197,127}#1}
\newcommand{\deitbest}[1]{\cellcolor[RGB]{96,174,92}#1}

\newcommand{\swinone}[1]{\cellcolor[RGB]{252,232,237}#1}
\newcommand{\swintwo}[1]{\cellcolor[RGB]{246,207,217}#1}
\newcommand{\swinthree}[1]{\cellcolor[RGB]{238,181,196}#1}
\newcommand{\swinfour}[1]{\cellcolor[RGB]{228,156,176}#1}
\newcommand{\swinbest}[1]{\cellcolor[RGB]{214,128,153}#1}

\def\BibTeX{{\rm B\kern-.05em{\sc i\kern-.025em b}\kern-.08em
    T\kern-.1667em\lower.7ex\hbox{E}\kern-.125emX}}
\begin{document}

\title{Adaptive Hebbian Memory Routing in Vision Transformers for Few-Shot Learning\\
}

\author{
\IEEEauthorblockN{
Mohammed Yusuf Mujawar, 
Noorbakhsh Amiri Golilarz\
}
\IEEEauthorblockA{
Department of Computer Science\\
The University of Alabama, Tuscaloosa, AL, USA\\
}
}

\maketitle

\begin{abstract}
Few-shot image recognition requires models to adapt to new classes from a small labeled support set. Hebbian fast-weight memory can provide temporary associative information during an episode, but fixed memory behavior may not be appropriate for every few-shot task. In this work, we propose Adaptive Hebbian Routing for few-shot Vision Transformers. The method uses a lightweight MLP router to control the contribution of Hebbian memory, the strength of memory updates, and the retention of previous memory from support-set features. We study Adaptive Placement, Adaptive Plasticity, and Fully Adaptive Hebbian Routing. Experiments use ViT-Small, DeiT-Small, and Swin-Tiny under 5-way 1-shot evaluation on Omniglot, CIFAR-FS, and cross-domain transfer from CIFAR-FS to Omniglot. In the direct Swin comparison, fixed and adaptive Hebbian variants use the same memory location. Adaptive Plasticity improves the fixed Hebbian result from 96.74\% to 96.92\%, while Fully Adaptive Routing achieves the best result at 96.94\%. The fully adaptive Swin model also reduces inference time from 16.51 ms to 14.05 ms relative to fixed Hebbian Swin. On CIFAR-FS, adaptive variants improve performance across all three backbones, and the multi-shot evaluation shows that these gains remain useful as the number of support examples increases. These results show that adaptive plasticity and adaptive memory activation can improve few-shot Transformer representations beyond fixed Hebbian behavior.
\end{abstract}


\section{Introduction}

Few-shot image recognition requires a model to recognize new classes
from only a small labeled support set. This setting is difficult because
standard deep networks store most of their knowledge in slow weights
that remain fixed during inference. Metric-based methods reduce this
difficulty by comparing query examples with the small support set. For
example, Prototypical Networks classify a query according to its
distance from class prototypes formed from support embeddings
\cite{snell2017prototypical}. However, the success of this process still
depends on whether the backbone can form useful task-specific
representations from the current episode.

Vision Transformers provide strong image representations, but they are
mostly static during inference. ViT represents images as sequences of
patch tokens \cite{dosovitskiy2021vit}, DeiT improves data-efficient
Transformer training \cite{touvron2021deit}, and Swin Transformer uses
a hierarchical shifted-window design for visual recognition
\cite{liu2021swin}. These architectures are effective feature
extractors, but few-shot learning requires them to adapt to a new
support set for every episode. This motivates mechanisms that can store
temporary task-specific information without updating the full backbone.

Hebbian fast weights provide one way to add this temporary memory.
Fast weights can store recent information on a shorter time scale than
standard model parameters \cite{hinton1987fast,ba2016fast}, while
Hebbian learning forms associations between co-activated
representations \cite{hebb1949organization}. This idea has been used
for meta-learning and one-shot binding \cite{munkhdalai2018hebbian},
and fast-weight formulations have also been connected to attention
mechanisms \cite{schlag2021linear}. These studies suggest that
associative memory can help a model adapt within an episode by storing
information from the support set and using it when processing queries.

However, the usefulness of Hebbian memory depends on how it is added to
the backbone. A recent study of Hebbian Fast-Weight modules in few-shot
Vision Transformers showed that fixed memory placement can produce very
different effects across architectures \cite{money2026where}. In
particular, fixed Hebbian memory was effective for Swin, while dense
per-block Hebbian insertion degraded ViT and DeiT. This result suggests
that selecting a useful memory location is important, but it also raises
a second question: once a location is selected, should the Hebbian
pathway use the same activation, update strength, and memory retention
for every few-shot episode?

In this work, we propose \textit{Adaptive Hebbian Routing}, a
lightweight approach that makes the Hebbian memory pathway
episode-conditioned. As illustrated in Fig.~\ref{fig:adaptive_router},
an adaptive MLP router receives support-set features from a selected
backbone location and predicts three control values. The first controls
how strongly the Hebbian memory output contributes to the current
representation. The second controls the plasticity rate used to write
new support-set associations. The third controls memory retention during
the episode. This design allows the model to change how it uses
temporary associative memory according to the current support set,
rather than applying fixed Hebbian behavior to every task.



The main contributions of this paper are as follows:
\begin{itemize}
    \item We propose Adaptive Hebbian Routing, which uses a lightweight
    MLP router to control Hebbian memory activation, plasticity strength,
    and memory retention from support-set features.

    \item We study Adaptive Placement, Adaptive Plasticity, and Fully
    Adaptive Hebbian Routing to separate the effect of memory activation
    from episode-specific memory updates.

    \item We provide a direct Swin comparison in which fixed and
    adaptive Hebbian variants use the same memory location, isolating
    the effect of adaptive control.

    \item We show that Adaptive Plasticity improves fixed Hebbian Swin,
    while Fully Adaptive Hebbian Routing achieves the strongest Swin
    accuracy and improves inference speed.

    \item We evaluate accuracy, parameter count, and inference time
    across Omniglot, CIFAR-FS, and cross-domain transfer from CIFAR-FS
    to Omniglot.
\end{itemize}

The remainder of this paper is organized as follows. Section~II reviews
related work on few-shot learning, Hebbian fast weights, and adaptive
routing. Section~III presents the proposed Adaptive Hebbian Routing
method and describes its integration into Vision Transformer backbones.
Section~IV describes the datasets, evaluation protocol, model variants,
and implementation details. Section~V reports the experimental results
and analysis. Section~VI discusses the main findings and limitations,
and Section~VII concludes the paper.

\section{Related Work}

\subsection{Few-Shot Learning with Learned Representations}

Few-shot learning aims to classify novel categories from a small labeled
support set. Early metric-based methods such as Matching Networks learn
to compare a query example with labeled support examples directly
\cite{vinyals2016matching}. Prototypical Networks simplify this setting
by representing each class with a prototype computed from its support
embeddings and assigning queries according to distance from those
prototypes \cite{snell2017prototypical}. Relation Networks instead learn
a comparison function between support and query representations
\cite{sung2018relation}. Other approaches focus on rapid parameter
adaptation, including Model-Agnostic Meta-Learning (MAML)
\cite{finn2017maml}, or on adapting the embedding space to each task
using set-to-set functions \cite{ye2020feat}. Our work follows the
episodic prototype-based setting, but focuses on improving the backbone
representation through temporary adaptive memory rather than changing
the classifier itself.

\subsection{Fast Weights and Hebbian Plasticity}

Fast-weight models separate long-term knowledge stored in slow
parameters from temporary associations formed during processing. Early
work described fast weights as a rapidly changing memory overlay that
can support temporary learning and binding \cite{hinton1987fast}. Later
work used fast weights to attend to recent information
\cite{ba2016fast} and showed that Hebbian fast weights can bind
representations and labels within a meta-learning task
\cite{munkhdalai2018hebbian}. The connection between attention and
fast-weight programming was further developed by work showing that
key-value outer-product updates can be interpreted as a fast
associative memory \cite{schlag2021linear}. Differentiable plasticity
also demonstrated that plastic connections can be optimized jointly
with neural network weights through gradient-based training
\cite{miconi2018differentiable}.

The fixed Hebbian Transformer baseline most closely related to our work
applied Hebbian Fast-Weight modules to Vision Transformer backbones for
few-shot character recognition \cite{money2026where}. Its results
showed that memory placement is architecture-dependent: fixed Hebbian
memory was effective for Swin, while dense per-block insertion was
harmful for ViT and DeiT. Our work builds on this finding by making the
Hebbian pathway episode-conditioned. In the direct Swin comparison,
fixed and adaptive Hebbian variants use the same memory location. This
allows us to evaluate whether Adaptive Placement and adaptive plasticity
improve memory behavior without changing where the Hebbian module is
inserted.

\subsection{Adaptive Routing and Conditional Computation}

Adaptive computation methods select or scale parts of a neural network
according to the input. Conditional computation uses learned activation
decisions to reduce unnecessary computation while preserving prediction
quality \cite{bengio2016conditional}. Adaptive Neural Networks and
BlockDrop select efficient execution paths for different inputs
\cite{bolukbasi2017adaptive,wu2018blockdrop}. For Vision Transformers,
DynamicViT predicts token importance and progressively removes less
informative tokens \cite{rao2021dynamicvit}. Routing has also been used
in mixture-of-experts models to direct examples or tokens to selected
expert subnetworks \cite{shazeer2017moe,fedus2022switch,zhou2022expert},
as well as in routing networks that select nonlinear functions for
different tasks \cite{rosenbaum2017routing}.

These methods mainly route computation through model blocks, tokens, or
experts. In contrast, Adaptive Hebbian Routing controls a temporary
associative memory pathway. The router does not replace the backbone or
choose a separate expert network. Instead, it determines how strongly
Hebbian memory should affect the current representation and how that
memory should be updated and retained during a few-shot episode.

\section{Method}

\subsection{Episodic Few-Shot Classification}

We consider an $N$-way $K$-shot classification episode with a labeled
support set
\begin{equation}
\mathcal{S} = \{(x_i, y_i)\}_{i=1}^{NK},
\end{equation}
and an unlabeled query set
\begin{equation}
\mathcal{Q} = \{x_j\}
\end{equation}
A backbone encoder maps support and query images into feature embeddings.
For each class $c$, a prototype is computed by averaging the embeddings
of its support examples:
\begin{equation}
p_c = \frac{1}{|\mathcal{S}_c|}
\sum_{(x_i,y_i)\in \mathcal{S}_c} f_{\theta}(x_i),
\end{equation}
where, $\mathcal{S}_c$ denotes the support examples belonging to class
$c$. A query image is classified according to its distance from the
class prototypes. We use squared Euclidean distance and optimize the
episodic classification loss over the query set.

Our method changes the feature encoder $f_{\theta}$ rather than the
prototype classifier. Specifically, we add adaptive Hebbian memory
pathways at selected locations in the Transformer backbone.

\subsection{Hebbian Fast-Weight Memory}

Each Hebbian module receives a token representation
\begin{equation}
X \in \mathbb{R}^{B \times T \times D},
\end{equation}
where, $B$ is the image batch size, $T$ is the token count, and $D$ is
the embedding dimension. The input is projected into key, value, and
query representations:
\begin{equation}
K = W_K X,\qquad V = W_V X,\qquad Q = W_Q X
\end{equation}

For each attention head, the module forms an associative key-value
update:
\begin{equation}
A = \operatorname{clip}
\left(
\frac{K^{\top}V}{\sqrt{T}},
-\delta,
\delta
\right),
\end{equation}
where, $\delta$ limits unusually large association values. The
fast-weight memory matrix is updated as
\begin{equation}
M \leftarrow \lambda M + \eta A,
\end{equation}
where, $\eta$ is the plasticity rate and $\lambda$ is the memory decay
factor. After each update, the memory matrix is normalized using its
Frobenius norm to keep its magnitude stable.

The memory is read using the query representation:
\begin{equation}
R = QM
\end{equation}
The retrieved signal is then modulated with an internal learned feature
gate, normalized, and added to the Transformer representation through a
residual connection. The memory matrix is reset at the start of every
few-shot episode. It therefore stores temporary associations from the
current support set rather than information that persists across
episodes.

\subsection{Adaptive MLP Router}

Fixed Hebbian memory uses the same activation and update behavior for
every few-shot episode. We replace this assumption with an adaptive MLP
router. A separate router is attached to each selected Hebbian location.
The router receives the support-token representation at that location
and computes an episode-level summary:
\begin{equation}
z_l = \operatorname{Mean}(X_l^{\mathcal{S}}),
\end{equation}
where, $X_l^{\mathcal{S}}$ contains the support features after the
$l$-th selected backbone location. The mean is taken across support
images and tokens, producing one feature vector for the current
episode.

The router is a lightweight two-layer MLP:
\begin{equation}
h_l = \operatorname{GELU}
\left(
W_1 \operatorname{LN}(z_l) + b_1
\right),
\end{equation}
\begin{equation}
r_l = W_2 h_l + b_2
\end{equation}

Depending on the adaptive variant, the router maps $r_l$ to one or more
control values:
\begin{equation}
g_l = \sigma(r_l^{g}),
\end{equation}
\begin{equation}
\eta_l = \eta_{\max}\sigma(r_l^{\eta}),
\end{equation}
\begin{equation}
\lambda_l = \sigma(r_l^{\lambda}),
\end{equation}
where, $g_l$ is the outer activation gate for the memory pathway,
$\eta_l$ is the episode-specific plasticity rate, and $\lambda_l$ is
the episode-specific memory decay factor. We use
$\eta_{\max}=0.15$ to keep the Hebbian update bounded.

The routed memory update becomes
\begin{equation}
M_l \leftarrow \lambda_l M_l + \eta_l A_l,
\end{equation}
and the memory-enhanced representation is
\begin{equation}
\widetilde{X}_l =
X_l + g_l \cdot H_l(X_l,M_l),
\end{equation}
where, $H_l(\cdot)$ denotes the Hebbian memory readout.

The support set writes a shared memory matrix, and both support and
query features read from that memory. Thus, the support examples form
an episode-specific associative memory that can influence query
representations before prototype classification.

\begin{figure}[!t]
\centering
\includegraphics[
    width=\columnwidth,
    height=0.62\textheight,
    keepaspectratio
]{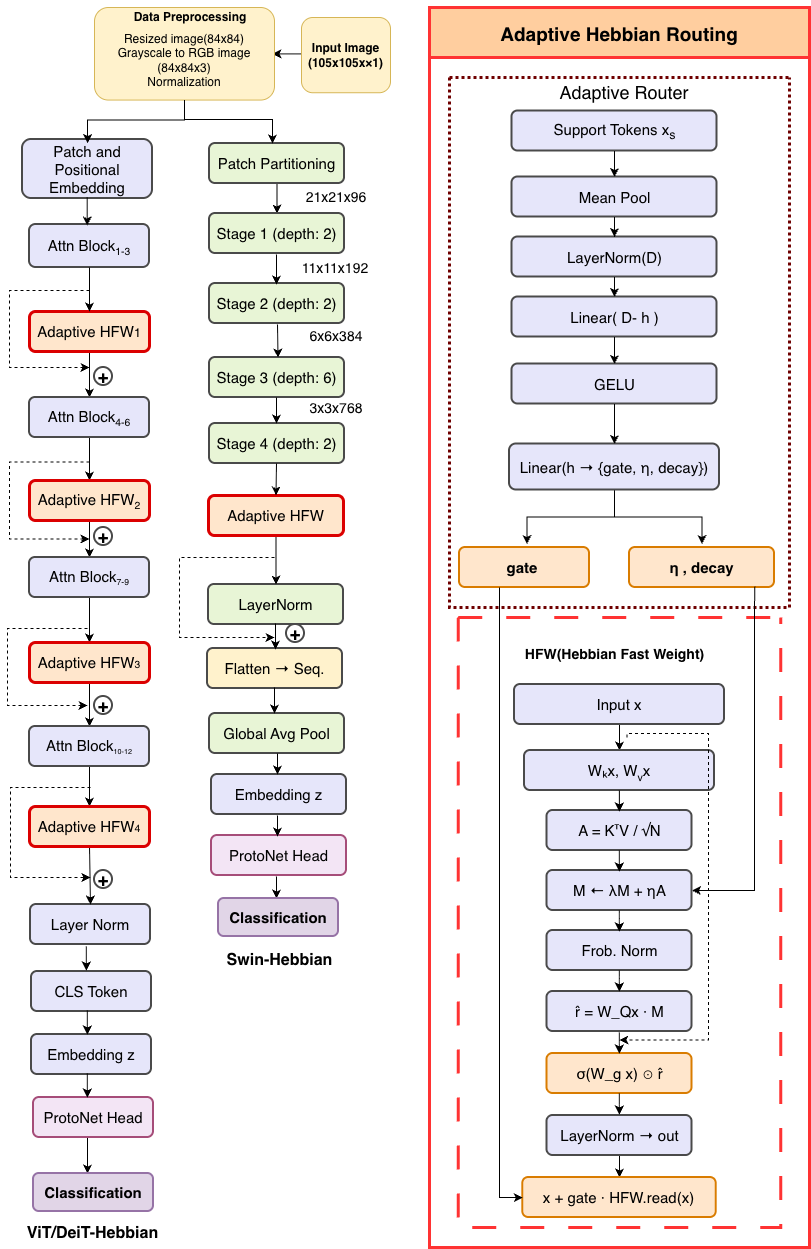}
\caption{Adaptive Hebbian Routing with an MLP router controlling memory contribution, plasticity, and decay.}
\label{fig:adaptive_router}
\end{figure}

Fig.~\ref{fig:adaptive_router} provides an overview of Adaptive
Hebbian Routing. The left side shows the integration of adaptive
Hebbian fast-weight modules within the ViT/DeiT and Swin backbones. The
router receives support-token features from a selected backbone
location, summarizes them through mean pooling, and predicts the memory
contribution gate, plasticity rate, and memory decay. These
episode-specific controls modulate the Hebbian fast-weight update and
the residual memory readout before feature extraction continues through
the backbone.

The router is intentionally lightweight. Its hidden dimension is set to
one eighth of the input embedding dimension, with a minimum of 16
hidden units. This keeps the routing overhead small relative to the
backbone and fixed Hebbian alternatives. The router output layers are
initialized near a conservative memory setting, with a low initial
outer gate, moderate plasticity, and high decay retention. This allows
the model to learn stronger or weaker memory use during training rather
than forcing large Hebbian updates at initialization.

\subsection{Adaptive Variants}

We study three variants to separate the effects of adaptive memory
activation and adaptive plasticity.

\noindent\textbf{Adaptive Placement.}
This variant learns the outer activation gate $g_l$, while the Hebbian
module retains its standard learned plasticity and decay parameters. It
tests whether the model benefits from learning how strongly the memory
pathway should contribute at each selected location.

\noindent\textbf{Adaptive Plasticity.}
This variant keeps the outer memory pathway fully active and uses the
router to predict $\eta_l$ and $\lambda_l$. It tests whether
episode-specific writing strength and memory retention are sufficient to
improve fast-memory behavior.

\noindent\textbf{Fully Adaptive Hebbian Routing.}
This variant predicts all three routing variables: $g_l$, $\eta_l$, and
$\lambda_l$. It allows the model to learn both how strongly memory
should contribute and how that memory should be updated for the current
few-shot episode.

\subsection{Backbone Integration}

For ViT-Small and DeiT-Small, adaptive Hebbian modules are inserted
after Transformer blocks 3, 6, 9, and 12. These locations distribute
the memory pathway across shallow, intermediate, and deep
representations while using fewer Hebbian modules than dense per-block
placement.

For Swin-Tiny, fixed and adaptive Hebbian variants use the same memory
location in the direct comparison. This provides a controlled test of
whether adaptive placement and adaptive plasticity improve a fixed
Hebbian design without changing where the memory module is inserted.
Adaptive Placement controls the contribution of the memory pathway,
Adaptive Plasticity controls its update strength and decay, and Fully
Adaptive Hebbian Routing controls all three variables.

At each selected location, support features first write the shared
fast-weight memory. Support and query features then read from the same
memory before continuing through the backbone. The memory is reset at
the beginning of every episode.

\section{Experimental Setup}

\subsection{Datasets and Evaluation Protocol}

We evaluate Adaptive Hebbian Routing on Omniglot and CIFAR-FS using
the episodic few-shot classification protocol. We report 5-way 1-shot episodic results, where each episode contains five novel classes with one labeled support example per class., where each episode contains five novel classes with one labeled support example per class. Query examples are classified using class prototypes computed from the support embeddings. We report mean episodic accuracy over the test episodes. We additionally evaluate 5-way CIFAR-FS performance across 1-, 3-,
5-, and 10-shot support sets to examine how the adaptive variants
behave as additional labeled examples become available.

For Omniglot, models are trained from scratch and evaluated on unseen
character classes. This setting allows a direct comparison with the
fixed Hebbian baseline under the same few-shot character recognition
protocol. For CIFAR-FS, ImageNet-pretrained ViT-Small, DeiT-Small, and
Swin-Tiny backbones are fine-tuned using episodic training on the
CIFAR-FS training classes. We also evaluate cross-domain transfer from
CIFAR-FS to Omniglot by testing the CIFAR-FS-trained models on Omniglot
without additional target-domain fine-tuning.

\subsection{Models and Adaptive Variants}

We evaluate ViT-Small, DeiT-Small, and Swin-Tiny backbones. For each
backbone, we compare the base model, the fixed Hebbian baseline, and
the proposed adaptive variants. The fixed baseline uses Hebbian
fast-weight memory with fixed activation, plasticity, and decay
behavior. Adaptive Placement predicts the contribution of the Hebbian
memory pathway while retaining standard learned plasticity parameters.
Adaptive Plasticity predicts the plasticity rate and memory decay while
keeping the memory pathway active. Fully Adaptive Hebbian Routing
predicts the memory contribution, plasticity rate, and memory decay
jointly.

For ViT-Small and DeiT-Small, the adaptive configuration uses four
selected Hebbian memory locations, whereas the fixed baseline uses a
dense Hebbian configuration across Transformer blocks. For Swin-Tiny,
the fixed and adaptive variants used in the direct comparison use the
same memory location. This allows the Swin evaluation to isolate the
effect of adaptive memory control from the effect of changing the
memory location.

\subsection{Training and Implementation Details}

All models are trained with episodic optimization using the query-set
classification loss. For Omniglot, the models are trained from scratch.
For CIFAR-FS, ImageNet-pretrained backbones are fine-tuned on the
CIFAR-FS training split before evaluation on the CIFAR-FS test split
and cross-domain transfer from CIFAR-FS to Omniglot. The fast-weight
memory matrix is reset at the beginning of every episode so that it
stores only temporary support-set associations.

All experiments are implemented in PyTorch and executed on an NVIDIA
RTX 5090 GPU. We report parameter count and inference time together
with episodic accuracy. Inference time is measured per few-shot episode
under the same implementation environment for the base, fixed Hebbian,
and adaptive variants.

\section{Results and Analysis}

For compact presentation, \textit{Fixed}, \textit{Place.},
\textit{Plast.}, and \textit{Full} denote Fixed Hebbian, Adaptive
Placement, Adaptive Plasticity, and Fully Adaptive Hebbian Routing,
respectively. All accuracy values are mean episodic test accuracy. In
Tables~\ref{tab:omniglot_results}--\ref{tab:cross_domain_results},
blue, green, and pink identify ViT, DeiT, and Swin, respectively.
Within each backbone row, color intensity increases from the lowest to
the highest accuracy. In Tables~\ref{tab:efficiency_omniglot} and
\ref{tab:efficiency_cifar}, darker shading indicates higher accuracy,
fewer parameters, or lower inference time within the corresponding
backbone family.

\subsection{Omniglot Results}

Table~\ref{tab:omniglot_results} compares the five variants when
trained from scratch on Omniglot. For ViT and DeiT, dense fixed Hebbian
insertion is harmful. Both base models reach 95.42\% accuracy, whereas
fixed Hebbian memory drops to 89.99\%. Adaptive Placement and Adaptive
Plasticity recover performance to 94.90\% and 94.92\%, respectively,
while Fully Adaptive Routing reaches 95.48\%. These results show that
episode-conditioned memory control removes most of the degradation
caused by dense fixed Hebbian insertion in flat Transformer backbones.

The Swin comparison provides the main direct evaluation of adaptive
memory control. Fixed and adaptive Hebbian variants use the same memory
location, so the comparison isolates the effect of changing memory
behavior rather than changing the location of the Hebbian module. Fixed
Hebbian memory improves the base Swin model from 96.14\% to 96.74\%.
Adaptive Placement reaches 96.63\%, which remains above the base model
but does not exceed fixed Hebbian memory. In contrast, Adaptive
Plasticity reaches 96.92\%, improving fixed Hebbian Swin by 0.18
percentage points. Fully Adaptive Routing obtains the strongest result
at 96.94\%, a 0.20 percentage-point improvement over fixed Hebbian
memory.

This result shows that adaptive placement alone is not sufficient in
this setting. The main improvement comes from episode-specific
plasticity and memory retention, while the combination of adaptive
placement and adaptive plasticity gives the best overall accuracy.

\begin{table}[!t]
\caption{Omniglot 5-way 1-shot accuracy (\%).}
\label{tab:omniglot_results}
\centering
\setlength{\tabcolsep}{2.0pt}
\renewcommand{\arraystretch}{1.10}
\begin{tabularx}{\columnwidth}{@{}lYYYYY@{}}
\toprule
Model & Base & Fixed & Place. & Plast. & Full \\
\midrule
ViT-Small
& \vitfour{95.42}
& \vitone{89.99}
& \vittwo{94.90}
& \vitthree{94.92}
& \vitbest{95.48} \\

DeiT-Small
& \deitfour{95.42}
& \deitone{89.99}
& \deittwo{94.90}
& \deitthree{94.92}
& \deitbest{95.48} \\

Swin-Tiny
& \swinone{96.14}
& \swinthree{96.74}
& \swintwo{96.63}
& \swinfour{96.92}
& \swinbest{96.94} \\
\bottomrule
\end{tabularx}
\end{table}

\subsection{CIFAR-FS Results}

Table~\ref{tab:cifar_results} reports results after fine-tuning
ImageNet-pretrained backbones on CIFAR-FS. This setting provides strong
evidence that adaptive memory control can improve few-shot adaptation.
For ViT, the base model reaches 69.58\% accuracy and fixed Hebbian
memory improves modestly to 72.58\%. Adaptive Placement, Adaptive
Plasticity, and Fully Adaptive Routing reach 87.55\%, 87.21\%, and
88.59\%, respectively. Fully Adaptive ViT improves over fixed Hebbian
memory by 16.01 percentage points.

DeiT follows the same general pattern. The base model reaches 60.56\%,
while fixed Hebbian memory reaches 80.48\%. Adaptive Placement achieves
the strongest DeiT result at 87.64\%, followed by Fully Adaptive
Routing at 86.50\%. Swin also benefits from adaptive memory control.
The base model reaches 52.52\%, fixed Hebbian memory reaches 74.03\%,
and Fully Adaptive Swin reaches 81.38\%. Across all three backbones,
the adaptive variants provide stronger CIFAR-FS performance than the
fixed Hebbian baseline.

\begin{table}[!t]
\caption{CIFAR-FS 5-way 1-shot accuracy (\%).}
\label{tab:cifar_results}
\centering
\setlength{\tabcolsep}{2.0pt}
\renewcommand{\arraystretch}{1.10}
\begin{tabularx}{\columnwidth}{@{}lYYYYY@{}}
\toprule
Model & Base & Fixed & Place. & Plast. & Full \\
\midrule
ViT-Small
& \vitone{69.58}
& \vittwo{72.58}
& \vitfour{87.55}
& \vitthree{87.21}
& \vitbest{88.59} \\

DeiT-Small
& \deitone{60.56}
& \deittwo{80.48}
& \deitbest{87.64}
& \deitthree{84.38}
& \deitfour{86.50} \\

Swin-Tiny
& \swinone{52.52}
& \swintwo{74.03}
& \swinfour{81.19}
& \swinthree{81.09}
& \swinbest{81.38} \\
\bottomrule
\end{tabularx}
\end{table}

\subsection{Performance Across Support-Set Sizes}

\begin{figure*}[!t]
\centering
\begin{tikzpicture}
\begin{groupplot}[
    group style={
        group size=3 by 1,
        horizontal sep=0.65cm
    },
    width=0.255\textwidth,
    height=0.245\textwidth,
    scale only axis,
    xlabel={Support shots},
    xmin=0.85, xmax=10.35,
    ymin=50, ymax=100,
    xtick={1,3,5,10},
    ytick={50,60,70,80,90,100},
    grid=major,
    grid style={dashed, gray!35},
    tick label style={font=\small},
    label style={font=\small},
    title style={font=\small},
    xlabel style={yshift=-2pt},
    clip=true,
    legend style={
        font=\small,
        draw=none,
        fill=none,
        at={(0.5,-0.40)},
        anchor=north,
        legend columns=5,
        column sep=4pt
    },
    every axis plot/.append style={
    line width=0.9pt,
    mark size=2.2pt
}
]

\nextgroupplot[
    title={ViT-Small},
    ylabel={Accuracy (\%)},
    legend to name=cifar_kshot_legend
]

\addplot[color=black, mark=*] coordinates {
    (1,69.81) (3,84.77) (5,88.04) (10,90.99)
};
\addlegendentry{Base}

\addplot[color=gray!75!black, mark=square*] coordinates {
    (1,71.59) (3,79.85) (5,83.09) (10,86.15)
};
\addlegendentry{Fixed}

\addplot[color=blue!75!black, mark=triangle*] coordinates {
    (1,87.63) (3,94.21) (5,93.76) (10,94.67)
};
\addlegendentry{Place.}

\addplot[color=orange!85!black, mark=diamond*] coordinates {
    (1,85.37) (3,92.15) (5,93.65) (10,94.73)
};
\addlegendentry{Plast.}

\addplot[color=red!75!black, mark=pentagon*] coordinates {
    (1,87.60) (3,93.47) (5,94.84) (10,95.47)
};
\addlegendentry{Full}

\nextgroupplot[
    title={DeiT-Small}
]

\addplot[color=black, mark=*] coordinates {
    (1,59.57) (3,85.92) (5,89.40) (10,93.01)
};

\addplot[color=gray!75!black, mark=square*] coordinates {
    (1,80.21) (3,86.89) (5,89.07) (10,91.31)
};

\addplot[color=blue!75!black, mark=triangle*] coordinates {
    (1,87.19) (3,93.75) (5,93.91) (10,94.56)
};

\addplot[color=orange!85!black, mark=diamond*] coordinates {
    (1,82.73) (3,90.16) (5,91.81) (10,93.33)
};

\addplot[color=red!75!black, mark=pentagon*] coordinates {
    (1,85.99) (3,92.71) (5,93.49) (10,94.63)
};

\nextgroupplot[
    title={Swin-Tiny}
]

\addplot[color=black, mark=*] coordinates {
    (1,54.00) (3,80.47) (5,86.57) (10,90.51)
};

\addplot[color=gray!75!black, mark=square*] coordinates {
    (1,73.64) (3,85.35) (5,88.67) (10,91.17)
};

\addplot[color=blue!75!black, mark=triangle*] coordinates {
    (1,80.93) (3,91.21) (5,91.29) (10,92.81)
};

\addplot[color=orange!85!black, mark=diamond*] coordinates {
    (1,79.65) (3,89.93) (5,92.35) (10,92.73)
};

\addplot[color=red!75!black, mark=pentagon*] coordinates {
    (1,82.64) (3,89.87) (5,92.21) (10,93.51)
};

\end{groupplot}
\end{tikzpicture}

\vspace{-0.5em}
\ref{cifar_kshot_legend}

\caption{CIFAR-FS accuracy across support-set sizes.}
\label{fig:cifar_kshot}
\end{figure*}
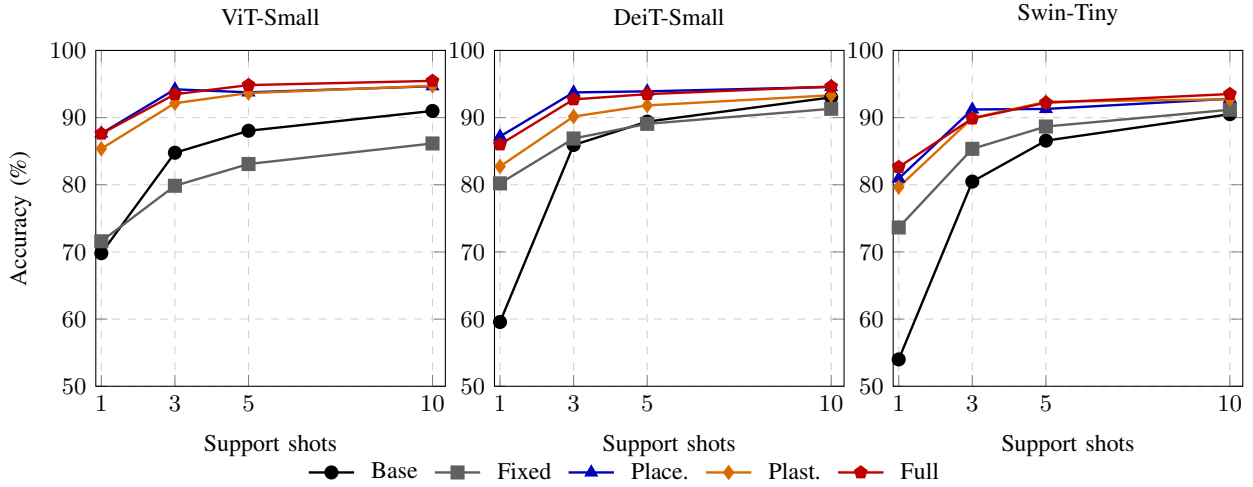

Fig.~\ref{fig:cifar_kshot} reports CIFAR-FS performance as the number
of support examples increases from 1-shot to 10-shot. All models improve
as more labeled support examples become available. The adaptive variants
remain strong across the support-set sizes. Fully Adaptive ViT reaches
95.47\% at 10-shot, while Fully Adaptive Swin reaches 93.51\%. These
results show that the benefit of adaptive Hebbian memory is not limited
to the 1-shot setting and remains useful when additional support
information is available.

The multi-shot curves are obtained from dedicated $K$-shot evaluation
runs. Therefore, the 1-shot values in Fig.~\ref{fig:cifar_kshot} may
differ slightly from the separately reported 1-shot results in
Table~\ref{tab:cifar_results}.

\begin{table}[!t]
\caption{Cross-domain transfer from CIFAR-FS to Omniglot: 5-way 1-shot accuracy (\%).}
\label{tab:cross_domain_results}
\centering
\setlength{\tabcolsep}{2.0pt}
\renewcommand{\arraystretch}{1.10}
\begin{tabularx}{\columnwidth}{@{}lYYYYY@{}}
\toprule
Model & Base & Fixed & Place. & Plast. & Full \\
\midrule
ViT-Small
& \vitone{54.35}
& \vitbest{58.46}
& \vitfour{56.77}
& \vittwo{54.75}
& \vitthree{56.57} \\

DeiT-Small
& \deitthree{74.50}
& \deittwo{72.84}
& \deitbest{77.03}
& \deitfour{75.10}
& \deitone{72.21} \\

Swin-Tiny
& \swintwo{72.28}
& \swinone{71.38}
& \swinthree{72.36}
& \swinfour{73.61}
& \swinbest{74.30} \\
\bottomrule
\end{tabularx}
\end{table}

\subsection{Cross-Domain Transfer from CIFAR-FS to Omniglot}

Table~\ref{tab:cross_domain_results} evaluates models fine-tuned on
CIFAR-FS and then tested on Omniglot without additional target-domain
fine-tuning. The results are mixed across backbones and provide a useful
test of how adaptive memory behaves under domain shift.

For Swin, Fully Adaptive Routing obtains the strongest cross-domain
result at 74.30\%, compared with 72.28\% for the base model and 71.38\%
for fixed Hebbian memory. Adaptive Plasticity also improves performance
to 73.61\%. For DeiT, Adaptive Placement achieves the strongest result
at 77.03\%, exceeding both the 74.50\% base model and the 72.84\% fixed
Hebbian model. These results suggest that adaptive memory behavior can
improve transfer when the source and target visual domains differ.

ViT shows a different pattern. Fixed Hebbian memory produces the
strongest ViT transfer result at 58.46\%, compared with 54.35\% for the
base model. Adaptive Placement and Fully Adaptive Routing improve over
the base model, but do not exceed fixed Hebbian memory, however they provide this transfer behavior with the same efficient adaptive inference design reported in the efficiency
analysis.

\subsection{Parameter and Inference Efficiency}

Tables~\ref{tab:efficiency_omniglot} and
\ref{tab:efficiency_cifar} compare accuracy, parameter count, and
inference time for ViT, DeiT, and Swin. The ViT and DeiT adaptive
variants use four Hebbian locations, whereas the fixed Hebbian baseline
uses dense Hebbian insertion across the Transformer blocks. On
Omniglot, this reduces the model size from 28.69M to approximately
24.04M parameters. A similar reduction is observed on CIFAR-FS, where
the adaptive ViT and DeiT models use approximately 24.06M parameters
compared with 28.71M for their fixed Hebbian counterparts.

The adaptive ViT and DeiT variants are also substantially faster than
the fixed Hebbian models. On Omniglot, Fully Adaptive ViT requires
10.31 ms per episode compared with 17.47 ms for fixed Hebbian ViT,
while Fully Adaptive DeiT requires 10.25 ms compared with 17.63 ms.
On CIFAR-FS, Fully Adaptive ViT reduces inference time from 24.51 ms
to 11.50 ms, and Fully Adaptive DeiT reduces inference time from
21.06 ms to 11.16 ms.

For Swin, the adaptive router introduces only a small parameter
overhead relative to fixed Hebbian memory. On Omniglot, Adaptive
Plasticity reaches 96.92\% accuracy with the fastest inference time of
13.83 ms. Fully Adaptive Routing reaches the best Swin accuracy,
96.94\%, with 14.05 ms inference time, compared with 16.51 ms for
fixed Hebbian Swin. On CIFAR-FS, Fully Adaptive Swin reaches 81.38\%
accuracy with 15.60 ms inference time, which is faster than fixed
Hebbian Swin at 17.69 ms.

\begin{table}[!t]
\caption{Omniglot efficiency comparison. Parameters are in millions; inference time is in milliseconds per episode.}
\label{tab:efficiency_omniglot}
\centering
\setlength{\tabcolsep}{1.6pt}
\renewcommand{\arraystretch}{1.08}
\begin{tabularx}{\columnwidth}{@{}llYYY@{}}
\toprule
Model & Variant & Accuracy & Params & \shortstack{Inference\\Time} \\
\midrule
ViT-Small
& Base   & \vitfour{95.42}  & \vitbest{21.60} & \vitbest{6.74} \\
& Fixed  & \vitone{89.99}   & \vitone{28.69}  & \vitone{17.47} \\
& Place. & \vittwo{94.90}   & \vitfour{24.04} & \vitfour{9.86} \\
& Plast. & \vitthree{94.92} & \vitfour{24.04} & \vitthree{10.26} \\
& Full   & \vitbest{95.48}  & \vitfour{24.04} & \vittwo{10.31} \\
\midrule
DeiT-Small
& Base   & \deitfour{95.42}  & \deitbest{21.60} & \deitbest{6.64} \\
& Fixed  & \deitone{89.99}   & \deitone{28.69}  & \deitone{17.63} \\
& Place. & \deittwo{94.90}   & \deitfour{24.04} & \deitfour{10.07} \\
& Plast. & \deitthree{94.92} & \deitfour{24.04} & \deittwo{10.34} \\
& Full   & \deitbest{95.48}  & \deitfour{24.04} & \deitthree{10.25} \\
\midrule
Swin-Tiny
& Base   & \swinone{96.14}   & \swinbest{27.51} & \swintwo{15.65} \\
& Fixed  & \swinthree{96.74} & \swintwo{29.87}  & \swinone{16.51} \\
& Place. & \swintwo{96.63}   & \swinone{29.94}  & \swinthree{14.17} \\
& Plast. & \swinfour{96.92}  & \swinone{29.94}  & \swinbest{13.83} \\
& Full   & \swinbest{96.94}  & \swinone{29.94}  & \swinfour{14.05} \\
\bottomrule
\end{tabularx}
\end{table}

\begin{table}[!t]
\caption{CIFAR-FS efficiency comparison. Parameters are in millions; inference time is in milliseconds per episode.}
\label{tab:efficiency_cifar}
\centering
\setlength{\tabcolsep}{1.6pt}
\renewcommand{\arraystretch}{1.08}
\begin{tabularx}{\columnwidth}{@{}llYYY@{}}
\toprule
Model & Variant & Accuracy & Params & \shortstack{Inference\\Time} \\
\midrule
ViT-Small
& Base   & \vitone{69.58}   & \vitbest{21.61} & \vitbest{7.49} \\
& Fixed  & \vittwo{72.58}   & \vitone{28.71}  & \vitone{24.51} \\
& Place. & \vitfour{87.55}  & \vitfour{24.06} & \vitthree{11.18} \\
& Plast. & \vitthree{87.21} & \vitfour{24.06} & \vitfour{11.17} \\
& Full   & \vitbest{88.59}  & \vitfour{24.06} & \vittwo{11.50} \\
\midrule
DeiT-Small
& Base   & \deitone{60.56}  & \deitbest{21.61} & \deitbest{7.30} \\
& Fixed  & \deittwo{80.48}  & \deitone{28.71}  & \deitone{21.06} \\
& Place. & \deitbest{87.64} & \deitfour{24.06} & \deittwo{11.44} \\
& Plast. & \deitthree{84.38}& \deitfour{24.06} & \deitthree{11.33} \\
& Full   & \deitfour{86.50} & \deitfour{24.06} & \deitfour{11.16} \\
\midrule
Swin-Tiny
& Base   & \swinone{52.52}   & \swinbest{27.51} & \swinthree{16.52} \\
& Fixed  & \swintwo{74.03}   & \swintwo{29.88}  & \swintwo{17.69} \\
& Place. & \swinfour{81.19}  & \swinone{30.10}  & \swinfour{15.70} \\
& Plast. & \swinthree{81.09} & \swinone{30.10}  & \swinone{23.72} \\
& Full   & \swinbest{81.38}  & \swinone{30.10}  & \swinbest{15.60} \\
\bottomrule
\end{tabularx}
\end{table}

\section{Discussion and Limitations}

The results show that the value of Hebbian memory depends not only on
where it is inserted, but also on how it is controlled during a
few-shot episode. Dense fixed Hebbian insertion is harmful for the flat
ViT and DeiT backbones on Omniglot. The adaptive variants recover most
of this performance loss, showing that a memory pathway should not
apply the same update behavior at every selected block and for every
task.

The direct Swin comparison gives the clearest evidence for adaptive
plasticity. Fixed Hebbian memory and all adaptive variants use the same
memory location. Adaptive Placement reaches 96.63\% and therefore does
not exceed the 96.74\% fixed-Hebbian result. However, Adaptive
Plasticity reaches 96.92\%, and Fully Adaptive Routing reaches 96.94\%.
Since all variants use the same insertion point, these improvements
cannot be explained by a placement change. They show that
episode-specific memory update strength and decay are important even
when the location of the memory module is already known to be useful.

The CIFAR-FS results provide complementary evidence. Adaptive routing
improves all three backbones relative to their base and fixed Hebbian
versions in this setting. For ViT and DeiT, the adaptive variants use
four selected Hebbian locations instead of dense fixed insertion across
the Transformer blocks. This reduces parameter overhead and inference
time while maintaining strong accuracy. The multi-shot CIFAR-FS results
further show that adaptive variants remain effective as the number of
support examples increases from 1-shot to 10-shot.

Cross-domain transfer remains backbone-dependent. Fully Adaptive Swin
and Adaptive Placement DeiT improve transfer from CIFAR-FS to Omniglot,
whereas fixed Hebbian memory remains strongest for ViT. This indicates
that adaptive memory control does not remove all effects of domain
shift. Instead, its usefulness depends on the source representation,
the target domain, and the backbone architecture.


\section{Conclusion}

We presented Adaptive Hebbian Routing, a lightweight approach for
controlling fast-weight memory in few-shot Vision Transformers. Rather
than applying fixed Hebbian behavior throughout a backbone, the method
uses an adaptive MLP router to control memory contribution, support-set
writing strength, and memory retention for each episode.

The direct Swin comparison shows the importance of adaptive plasticity.
When fixed and adaptive Hebbian variants use the same memory location,
Adaptive Plasticity improves fixed Hebbian accuracy from 96.74\% to
96.92\%, while Fully Adaptive Hebbian Routing achieves the strongest
result at 96.94\%. The fully adaptive Swin model also reduces inference
time relative to fixed Hebbian Swin. These findings show that the
benefit of Hebbian memory depends not only on using a suitable memory
location, but also on adapting how that memory is updated for the
current few-shot episode.

Across Omniglot, CIFAR-FS, and cross-domain transfer from CIFAR-FS to
Omniglot, adaptive memory control reduces the weakness of dense fixed
Hebbian insertion in ViT and DeiT while providing strong few-shot
adaptation results. The CIFAR-FS multi-shot evaluation further shows
that the adaptive variants remain effective as the number of support
examples increases. Overall, Adaptive Hebbian Routing provides a
selective and efficient way to use temporary associative memory by
learning how strongly it should contribute and how it should be updated.

\section*{Acknowledgment}

The authors acknowledge the support and resources provided by the
Bioinspired Robotics, AI, Imaging and Neurocognitive Systems (BRAINS)
Laboratory at The University of Alabama.

\bibliographystyle{IEEEtran}
\bibliography{references_adaptive_hebbian_final}

\end{document}